# *Some Epistemological Problems with the Knowledge Level in Cognitive Architectures*

*Antonio Lieto*
*Università di Torino, Dipartimento di Informatica, Italy*
*ICAR-CNR, Palermo, Italy*
*lieto@di.unito.it, lieto.antonio@gmail.com*

**Introduction**

This article addresses an open problem in the area of cognitive systems and architectures: namely the problem of handling (in terms of processing and reasoning capabilities) complex knowledge structures that can be at least plausibly comparable, both in terms of size and of typology of the encoded information, to the knowledge that humans process daily for executing everyday activities.
Handling a huge amount of knowledge, and selectively retrieve it according to the needs emerging in different situational scenarios, is an important aspect of human intelligence. For this task, in fact, humans adopt a wide range of heuristics (Gigerenzer & Todd) due to their "bounded rationality" (Simon, 1957). In this perspective, one of the requirements that should be considered for the design, the realization and the evaluation of intelligent cognitively-inspired systems should be represented by their ability of heuristically identify and retrieve, from the general knowledge stored in their artificial Long Term Memory (LTM), that one which is synthetically and contextually relevant. This requirement, however, is often neglected. Currently, artificial cognitive systems and architectures are not able, *de facto*, to deal with complex knowledge structures that can be even slightly comparable to the knowledge heuristically managed by humans. In this paper I will argue that this is not only a technological problem but also an epistemological one and I will briefly sketch a proposal for a possible solution.

**1.1 The Knowledge Problem**

As mentioned, current cognitive artificial systems and cognitive architectures are not equipped with knowledge bases comparable with the conceptual knowledge that humans possess and use in the everyday life. From an epistemological perspective this lack represents a problem: in fact, endowing cognitive agents with more "realistic" knowledge bases, in terms of both the **size** and the **type of information** encoded, would allow, at least in principle, to test the different artificial systems in situations closer to that one encountered by humans in real-life.
This problem becomes more relevant if we take into account the Cognitive Architectures (Newell, 1990). While "cognitively-inspired systems", in fact, could be designed to deal with only domain-specific information (e.g. let us think to a computer simulator of a poker player), Cognitive

Architectures, on the other hand, have also the goal and the general objective of testing - computationally - the general models of mind they implement.

Therefore: if such architectures only process a simplistic amount (and a limited typology) of knowledge, the structural mechanisms that they implement concerning knowledge processing tasks (e.g. that ones of retrieval, learning, reasoning etc.) can be only loosely evaluated, and compared w.r.t. that ones used by humans in similar knowledge-intensive tasks. In other words: from an epistemological perspective, the explanatory power of their computational simulation is strongly affected (Minkowski, 2013).

**1.2 The "Knowledge Limit" of Cognitive Architectures**

The design and adaptation of cognitive architectures is a wide and active area of research in Cognitive Science, Artificial Intelligence and, more recently, in the area of Computational Neuroscience. Cognitive architectures have been historically introduced i) "to capture, at the computational level, the invariant mechanisms of human cognition, including those underlying the functions of control, learning, memory, adaptivity, perception and action" (Oltramari and Lebiere, 2012) and ii) to reach human level intelligence, also called AGI (Artificial General Intelligence), by means of the realization of artificial artifacts built upon them. During the last decades many cognitive architectures have been realized, - such as ACT-R (Andersson et al. 2004), SOAR (Laird 2008) etc. - and have been widely tested in several cognitive tasks involving learning, reasoning, selective attention, recognition etc. However, they are general structures without a general content.

Thus, every evaluation of systems relying upon them is necessarily task-specific and do not involve not even the minimum part of the full spectrum of processes involved in the human cognition when the "knowledge" comes to play a role. This means that the knowledge embedded in such architectures is usually ad-hoc built, domain specific, or based on the particular tasks they have to deal with. Such limitation, however, do not provide any advancement in the cognitive research about how the humans heuristically select and deal with the huge and variegated amount of knowledge they possess when they have to: make decisions, reason about a given situation or, more in general, solve a particular cognitive task involving several dimensions of analysis.

This problem, as a consequence, also limits the advancement of the research in the area of Artificial General Intelligence. The "knowledge" limit of the cognitive architectures has been recently pointed out in literature (Oltramari and Lebiere 2012) and some technical solutions for filling this "knowledge gap" has been proposed. In particular the use of ontologies and of semantic formalisms has been seen as a possible solution for providing effective content to the structural knowledge modules of the cognitive architectures. Some initial efforts have been done in this sense. In particular, within the MindEye project, the ACT-R architecture developed at the Carnegie Mellon University has been "semantically extended" with the ontological content coming from three integrated semantic resources composed by the lexical databases WordNet (Fellbaum 1998), FrameNet and by a branch of the top level ontology DOLCE (Masolo et al. 2003) related to the event modelling. However, also in this case, the amount of semantic knowledge selected for the real-ization of the Cognitive Engine (one of the systems developed within the MindEye Program), and for its evaluation, was tailored on the specific needs of the system itself. It, in fact, was aimed at solving a precise task of event recognition trough a video-surveillance intelligent machinery; therefore only the ontological knowledge about the "events" was selectively embedded in it.

## 2 A Distributed Approach: Heterogeneous Knowledge Frameworks in the Linked Data Ecosystem

Despite the importance of these first attempts aimed at connecting ontologies and cognitive architectures, this approach still present problems within the research program aimed at achieving an effective Artificial General Intelligence (AGI) since both in terms of "size" and in terms of "types of encoded knowledge" the problems mentioned above still persist. On the other hand, a viable solution able to promote an effective extension and adoption of semantic content within the cognitive architectures should go beyond the integration of standard symbolic knowledge represented in semantic format. In particular, differently from what has been already proposed, the solution defended in this paper does not suggest to connect single ontologies to the knowledge modules of the cognitive architectures. It is based, on the other hand, on the idea of connecting multiple and heterogeneous knowledge spaces and frameworks to the available knowledge modules of the cognitive architectures. The proposed approach is somehow related to a recent research trend, developed within the Semantic Web research community, known as Linked Data (Bizer et al. 2009).

Following this view, in recent years, a huge amount of semantic data (released in standard semantic web languages such as RDF) has been published on the web (as "knowledge bubbles" or "knowledge endpoints"), linked and integrated together with other "knowledge endpoint". The ultimate goal of such linkage has been that one of creating a unified semantic knowledge space available in a machine readable format. Famous examples of such interconnected "knowledge bubbles" are DBpedia (the semantic version of Wikipedia, in RDF).

The main technical advantage coming from this integration is represented by the possibility of using such linked knowledge as an effective alternative to the standard solution based on the equipment of cognitive architectures with monolithic pieces of, ad hoc, selected ontological knowledge used for solving specific problems. On the other hand, such knowledge space allows, at least in principle, the technical possibility of encoding, for the first time, a general linked knowledge within a general cognitive architecture.

The integration with the "world-level knowledge" (see Salvucci, 2014) by means of external knowledge sources is, however, a necessary but not sufficient condition for solving the problem of the "knowledge level" in Cognitive Architectures since it only deals with the "size" aspect. One of the main problems of the "knowledge-bubbles" encoded in the Linked Data perspective is, in fact, represented by its **homogeneity**: i.e. only one part of the whole spectrum of conceptual informations is encoded in this kind of symbolic based representations (usually the so called "classical" part: that one representing concepts in terms of necessary and sufficient information, see Frixione and Lieto, 2012 on these aspects). On the other hand, the so called "**common-sense**" conceptual components (i.e. those that, based on the notions developed in the experimental and theoretical cognitive sciences, allow to characterize concepts in terms of "prototypes", "exemplars" or "theories", see Frixione and Lieto, 2013; and Frixione and Lieto, 2014 for further details) is largely absent in such framework. Common-sense conceptual knowledge, however, is exactly the type of "cognitive information" crucially used by humans for heuristic reasoning and decision making.

Given this state of affairs, a viable solution for achieving an effective simulation of human-level conceptual representation and processing, would require to endow the knowledge ecosystem of the Linked Data approach with a **heterogeneous perspective** to the representation of conceptual knowledge. In particular: it would be necessary to introduce conceptual frameworks able to represent the information not only in symbolic and classical logic-oriented fashion but also in a common-sense one. A suitable solution for the representation (and reasoning) of common sense knowledge is represented by the Peter Gärdenfors's proposal of Conceptual Spaces (Gärdenfors, 2014). In such framework at least prototypical and exemplars-based representations (and their corresponding reasoning mechanisms) can be naturally represented. A further argument supporting this framework as a good candidate for an extended *heterogeneous* Linked Data ecosystem (LDe) is the existence of a XML-like Conceptual Space mark-up language (CSML) that could be easily extendable towards the RDF language (the standard adopted within the Linked Data community)[1].

Summing up: in this paper I have tried to briefly present (the brevity is due to the lack of space) two main epistemological problems affecting the knowledge level in the Cognitive Architectures. After detailing the problems I have suggested that, from a technical perspective, the adoption of a **hetereogeneous** representational approach, within the LDe, could allow to deal with both the "*size*" argument and with that one concerning the "*typology*" of the encoded knowledge, thus providing at least in principle, a possible solution concerning the current epistemological limitations of the knowledge level in cognitive architectures. Within the family of the approaches claiming for the need of representational heterogeneity for the conceptual knowledge, the idea of representing conceptual structures as "**heterogeneous proxytypes**" (see Lieto, 2014) seems to be particularly feasible for providing the integration between the heterogeneous representational level and the corresponding reasoning procedures embedded in general purpose Cognitive Architectures (due to the lack of space I remind the interested reader to the references indicated above and below for an introduction to the mentioned approach). The first systems designed according to such approach, and integrated with existing Cognitive Architectures, have obtained, in fact, encouraging results in task of conceptual categorization and retrieval if compared with humans answers (on these aspects see Lieto et al. 2015; Lieto et al. forthcoming). Additional investigations are, however, needed (and represent ongoing work) in order to deeply evaluate the efficacy of the proposed approach in more challenging scenarios.

---

[1] Of course, other frameworks oriented towards the representation of additional kinds of "common-sense" knowledge (e.g. that one hypothesized within the "theory-theory") could be introduced as well in order to enhance the variety of knowledge usable by a cognitive agent whose behaviour and processes are controlled within the framework of a Cognitive Architecture.